\documentclass[twoside,leqno,twocolumn]{article}  
\usepackage{ltexpprt}

\usepackage{amssymb}
\usepackage{amsmath}

\usepackage{graphicx}
\usepackage{multirow}

\usepackage{algorithm}
\usepackage{algorithmic}

\usepackage{textcomp}
\usepackage{balance}
\usepackage{url}

\newenvironment{noindlist}
 {\begin{list}{\labelitemi}{\leftmargin=0.5em \itemindent=0.5em}}
 {\end{list}}

\begin{document}

\title{\Large Discriminative Density-ratio Estimation}
\author{
Yun-Qian Miao    \thanks{Center of Pattern Analysis and Machine Intelligence, University of Waterloo, Waterloo, Canada N2L 3G1. (yqmiao@uwaterloo.ca)}
\and 
Ahmed K. Farahat \thanks{Center of Pattern Analysis and Machine Intelligence, University of Waterloo, Waterloo, Canada N2L 3G1. (afarahat@uwaterloo.ca)}
\and
Mohamed S. Kamel \thanks{Center of Pattern Analysis and Machine Intelligence, University of Waterloo, Waterloo, Canada N2L 3G1. (mkamel@uwaterloo.ca)}
}

\date{}

\maketitle


\begin{abstract}
\small\baselineskip=9pt The covariate shift is a challenging problem in supervised learning that results from the discrepancy between the training and test distributions. An effective approach which recently drew a considerable attention in the research community is to reweight the training samples to minimize that discrepancy. In specific, many methods are based on developing Density-ratio (DR) estimation techniques that apply to both regression and classification problems. Although these methods work well for regression problems, their performance on classification problems is not satisfactory. This is due to a  key observation that these methods focus on matching the sample marginal distributions without paying attention to preserving the separation between classes in the reweighted space. In this paper, we propose a novel method for Discriminative Density-ratio (DDR) estimation that addresses the aforementioned problem and aims at estimating the density-ratio of joint distributions in a class-wise manner. The proposed algorithm is an iterative procedure that alternates between estimating the class information for the test data and estimating new density ratio for each class.
To incorporate the estimated class information of the test data, a soft matching technique is proposed. In addition, we employ an effective criterion which adopts mutual information as an indicator to stop the iterative procedure while resulting in a decision boundary that lies in a sparse region. 
Experiments on synthetic and benchmark datasets demonstrate the superiority of the proposed method in terms of both accuracy and robustness.
\end{abstract}

\textbf{Keywords:} Covariate Shift; Density-ratio Estimation; Cost-sensitive Classification.

\section{Introduction}
\label{sec1}

There are many real world applications where the test data demonstrates distribution shift from the training data. In such cases, traditional machine learning techniques usually encounter performance degradation because their models are fitted towards minimizing error or risk of error for the training samples. So it is important that the learning algorithms can demonstrate some degree of \textit{adaptivity} to cope with distribution changes. This has resulted in intensive research under the names domain adaptation \cite{DaumeIII2006, Ben-David2010}, transfer learning \cite{Pan2010}, and concept drift \cite{Tsymbal2004, Bose2011}. One particular case is the covariate shift problem \cite{Shimodaira2000, Moreno-Torres2012}, which assumes that the marginal distributions are changed between training and test data (i.e., $ p_{ts}(x)\neq p_{tr}(x)$), while the class conditional distributions are not affected (i.e., $p_{ts}\left(y|x\right)=p_{tr}\left(y|x\right)$).

Usually the covariate shift happens in biased sample selection scenarios. For example, in building an action recognition system, the training samples are collected in a university lab, where young people make up a high percentage of the population. When the system is intended to be applied in reality, it is likely that we will face a more general population model.

To compensate for the distribution gap between the training and test data, the objective of model fitting is modified to minimize the expectation of the weighted error, where the weight of the training sample is justified using its density ratio \cite{Shimodaira2000}, i.e., $\beta(x) = p_{ts}(x)/p_{tr}(x)$. Therefore, the solution to covariate shift is formulated as estimating the marginal density ratio and applying some cost-sensitive learning techniques \cite{Zadrozny2003, Sun2007}.

Recently, a number of methods have been proposed to estimate the Density ratio (DR), given two sets of finite number of observation samples. There are two groups of methods for Density-ratio estimation in the literature. One is a two-step procedure: first estimate the training and test probability densities separately and then divide them. The second group of methods estimates the density ratio directly in one step. These one-shot methods usually achieve more accurate and robust results and are considered the state of the art~\cite{Huang2007, Gretton2009, Sugiyama2007, Yu2012}.

In the literature, the aforementioned reweighting of training samples according to the density ratio is examined in a wide range of applications, including both the regression and classification tasks. We have seen these reweighting methods performing well in many regression tasks. However, existing research and our experiments found that these methods do not yield satisfactory results in classification scenarios. In many cases, they even recorded worse prediction accuracy than the simple unweighted approach. This motivated us to investigate the covariate shift classification problems in this work. Our key observation is that these conventional density-ratio methods focus on matching the training and test distributions without paying attention to preserving the separation among classes in the reweighted space. So these traditional density-ratio estimation methods may deteriorate the discrimination ability even if the marginal distributions might be matched well in general.

In this paper, we propose a novel method called Discriminative Density-ratio (DDR) estimation that addresses the aforementioned problem and aims to estimate the density ratio between the joint distributions in a class-wise manner. To do so, we divide the task into two parts: (1) estimating the density ratio between the training and test data for each class; and (2) estimating the class prior changes for the test data. As the class labels for the test data are unknown, the proposed method is based on an iterative procedure, which alternates between estimating the class information for the test data and estimating new class-wise density ratios.

In comparison to the conventional approach which matches sample marginal distributions, the proposed class-wise matching method has two benefits. First, it allows relaxing the assumption of the covariate shift that $p_{ts}\left(y|x\right)=p_{tr}\left(y|x\right)$ and accordingly captures a mixture of distribution changes. Second, it focuses on the classification problems and considers preserving the separation among classes while matching the shifted distributions. Our experiments on synthetic and benchmark data confirm the effectiveness of the proposed DDR algorithm.

The paper is organized as follows: The rest of this section describes the notations used in the paper. Section \ref{rev} introduces the covariate shift problem. Section \ref{pre} reviews the state of the art on the density-ratio estimation and analyzes the limitations of previous work. Section \ref{ddr} describes our proposed method. In Section \ref{exp}, empirical evaluations are conducted. Section \ref{con} concludes the paper.

\subsection{Notation.}

Throughout this paper, scales, vectors, and matrices are shown in small, bold, and capital letters respectively. When discussing covariate shift classification, we use the following notations:

\begin{tabular} {p{1cm}p{6cm}}
 $\mathcal{X}$ & $\mathcal{X} \subseteq \mathbb{R}^{d}$, the $d$-dimension input space, $x\in\mathcal{X}$ is an input sample \\
 $\mathcal{Y}$ & the class label space, $y\in\mathcal{Y}$ is an output variable \\
\end{tabular}

\begin{tabular} {p{1cm}p{6cm}}
 $p_{tr}$ & the probability density of the training data \\
 $p_{ts}$ & the probability density of the test data \\
 $n_{tr}$ & the number of training samples \\
 $n_{ts}$ & the number of test samples \\
 $\pi_{tr}$ & the set of training samples \\
 $\pi_{ts}$ & the set of test samples \\
 $\beta$ & the density-ratio between two distributions\\
 $\gamma$ & the ratio between two class priors\\
\end{tabular}

\section{Learning Under Covariate Shift}
\label{rev}

With the empirical risk minimization framework \cite{Vapnik1998, Gretton2009}, the general purpose of a supervised learning problem is to minimize the expected risk of
\begin{equation}
\boldsymbol{\text{R}}\left(\theta,p,l\right)=\iint l\left(x,y,\theta\right)p\left(x,y\right)dxdy \:,
\end{equation}
where $\theta$ is a learned model, $l\left(x,y,\theta\right)$ is a loss function for the problem with a joint distribution $p\left(x,y\right)$.

When we are facing the case where the training distribution $p_{tr}(x,y)$ differs from the distribution of test data $p_{ts}(x,y)$, in order to obtain the optimal model in the test domain $\theta_{ts}^{*}$, we can derive the following reweighting scheme:
\begin{eqnarray}
\theta_{ts}^{*} & = & \arg\min_{\theta\in\boldsymbol{\theta}}R_{ts}\left[\theta,p_{ts}\left(x,y\right),l\left(x,y,\theta\right)\right]\nonumber \\
 & = & \arg\min_{\theta\in\boldsymbol{\theta}}\iint l\left(x,y,\theta\right)p_{ts}\left(x,y\right)dxdy\nonumber \\
 & = & \arg\min_{\theta\in\boldsymbol{\theta}}\iint l\left(x,y,\theta\right)\frac{p_{ts}\left(x,y\right)}{p_{tr}\left(x,y\right)}p_{tr}\left(x,y\right)dxdy\nonumber \\
 & = & \arg\min_{\theta\in\boldsymbol{\theta}}R_{tr}\left[\theta,p_{tr}\left(\boldsymbol{x,y}\right),\frac{p_{ts}\left(x,y\right)}{p_{tr}\left(x,y\right)}l\left(x,y,\theta\right)\right]\nonumber \\
 & \approx & \arg\min_{\theta\in\boldsymbol{\theta}}\frac{1}{n_{tr}}\sum_{(x,y)\in\pi_{tr}}\frac{p_{ts}\left(x,y\right)}{p_{tr}\left(x,y\right)}l\left(x,y,\theta\right) \:.\label{eq:rts}
\end{eqnarray}

Further, covariate shift assumes that the class conditional distributions are the same across the training and test data (i.e. $p_{ts}\left(y|x\right)=p_{tr}\left(y|x\right)$), but that the marginal distributions are different. Hence we can derive that:
\begin{eqnarray}
\theta_{ts}^{*} & \approx & \arg\min_{\theta\in\boldsymbol{\theta}}\frac{1}{n_{tr}}\sum_{(x,y)\in\pi_{tr}}\frac{p_{ts}\left(y|x\right)p_{ts}\left(x\right)}{p_{tr}\left(y|x\right)p_{tr}\left(x\right)}l\left(x,y,\theta\right)\nonumber \\
 & = & \arg\min_{\theta\in\boldsymbol{\theta}}\frac{1}{n_{tr}}\sum_{(x,y)\in\pi_{tr}}\frac{p_{ts}\left(x\right)}{p_{tr}\left(x\right)}l\left(x,y,\theta\right)\nonumber \\
 & = & \arg\min_{\theta\in\boldsymbol{\theta}}\frac{1}{n_{tr}}\sum_{(x,y)\in\pi_{tr}}\beta(x)l\left(x,y,\theta\right) \:.\label{eq:Rtsbeta}
\end{eqnarray}

Now, the learning objective in the new test domain would
be evaluated by the importance-weighted training samples to reflect
the changes of distribution, where the importance of a sample
is equal to the density-ratio $\boldsymbol{\beta}$. 

Having the weighted training instances, there are plenty of cost-sensitive
learning algorithms that can be applied. Instead of minimizing the
loss of mis-classification, the cost-sensitive learning aims at minimizing
the instance-dependent cost of wrong prediction~\cite{Elkan2001,Zadrozny2003,Sun2007}.
For example, the Support Vector Machines~\cite{Chang2011a} and Regularized
Least Squares~\cite{Sugiyama2007} can naturally embed weighted samples
in the training process.

\section{Density-ratio Estimation}
\label{pre}

Because of the increasing demand from practical application domains to develop machine learning systems that adapt to unseen cases, the Density-ratio (DR) estimation has attracted considerable attention in the research community, and there are numerous methods being proposed to solve the problem in the literature. The simple approach is to solve the density-ratio estimation problem in two steps: estimating the training and test distributions separately and taking a division. However, this na\"{i}ve method encounters several problems \cite{Huang2007}: 
\begin{enumerate}
\item The information from the given limited number of samples may be sufficient to infer the density-ratio, but insufficient to infer two probability density functions. The estimation of probability density is usually a more general and challenging problem.
\item A small estimation error in the denominator can lead to a large variance in the density-ratio.
\item The na\"{i}ve approach would be highly unreliable for high-dimension problems because of the well-known ``curse-of-dimensionality" problem. 
\end{enumerate}

Therefore, researchers have been putting efforts on proposing new methods to estimate the density-ratio directly without going through the estimation of two probability densities. Along this direction, Huang et al. proposed a Kernel Mean Matching (KMM) algorithm \cite{Huang2007}, which directly gives the estimates of sample importance by minimizing the mean discrepancy in a reproduced kernel space. By explicitly modeling the function of density-ratio, another group of methods have been developed
with the formulation of various objective functions, which include the Kullback-Leibler Importance Estimation Procedure (KLIEP) \cite{Sugiyama2007}, Least-Squares Importance Fitting (LSIF) \cite{Kanamori2008}, unconstrained Least-Squares Importance Fitting (uLSIF) \cite{Kanamori2008}. Among these density-ratio estimation methods, various advantages are demonstrated
on their different applicable fields. In general, the method of uLSIF was shown to have excellent numerical stability and efficient run-time solution \cite{Sugiyama2009}.

Besides the covariate shift problem mentioned above, the density-ratio
estimation has shown noticeable potential in many data mining and
machine learning fields. Some highlighted applicable fields are outliers
detection~\cite{Hido2011}, change-point detection for time series
data, feature selection and feature extraction
based on mutual information estimation~\cite{Sugiyama2009}.

\subsection{Limitations of Previous Work.}

Reviewing the existing work on the importance reweighting strategy for covariate shift adaptation, the first issue is the strong assumption on the class conditional distributions (the posterior). The posteriors are assumed to be fixed between the training and test data, while the marginal distributions exhibit changes (i.e., $p_{ts}\left(y|x\right)=p_{tr}\left(y|x\right)$ and $ p_{ts}(x)\neq p_{tr}(x)$). According to the Bayes' rule, there is the following equation
that describes the relationship between the prior, posterior, marginal, likelihood, and joint distributions as
\begin{equation}
 p(x,y)=p(y|x)p(x)=p(x|y)p(y) \:.
\end{equation}

The root cause of covariate shift is the sampling bias, i.e., $p_{ts}(x)\neq p_{tr}(x)$. In this giving condition, we can not assure the posterior will remain unchanged, even if we know that the concepts behind the data are stable. This means that when the distributions of covariate are shifted, there is a high possibility that the class conditional distributions and/or priors
will also change.

Moreover, focusing on the classification problem, the objective is to discriminatively separate the instances into different classes. However, in the conventional weighting approach dealing with covariate shift adaptation, the distribution
matching is performed on the whole input space. In other words, the existing algorithms focus on matching the training and test distributions without considering to preserve the separation among classes in the reweighted space.

These two problems heavily hold back the effectiveness of the weighting methods in correcting the covariate shift problem, especially for classification tasks. Several research reported this fact \cite{Gretton2009,Cortes2008,Sugiyama2009},
but none of them presented a clear solution. 

\section{Proposed Approach} 
\label{ddr}

Having the intuition of preserving the separations between classes while pursuing the match of distributions, we propose an approach named Discriminative Density-ratio (DDR) estimation, which aims to estimate the density ratio between the joint distributions in a class-wise manner. Our proposed approach uses an iterative procedure. The class labels of the test samples are estimated using the updated density-ratio estimates and in turn the density ratios are estimated for each class.

Following Eq. (\ref{eq:rts}), instead of assuming unchanged class conditional distributions and simplifying Eq. (\ref{eq:rts}) into the density ratio on $x$, we decompose the joint distributions from the perspective of class likelihood and define a more general weighting scheme $\boldsymbol{w}$ to reflect the density ratio of joint distributions as 
\begin{eqnarray}
\boldsymbol{w} & = & \frac{p_{ts}(x,y)}{p_{tr}(x,y)}\nonumber \\
           & = & \frac{p_{ts}(x|y)\cdot p_{ts}(y)}{p_{tr}(x|y)\cdot p_{tr}(y)}\:.
\end{eqnarray}

For a classification problem, assuming class labels are from a finite discrete set $y\in[c_{1},c_{2},\ldots,c_{m}]$, then the density ratio of joint distributions can be evaluated in a class-wise manner as
\begin{equation}
\boldsymbol{w} =  \left[{w}_{c_{1}}, {w}_{c_{2}}, \ldots, {w}_{c_{m}}\right]^{T} \:,
\end{equation} where
\begin{equation}
{w}_{c_i} =  \frac{p_{ts}(x|y=c_{i})}{p_{tr}(x|y=c_{i})}\cdot\frac{p_{ts}(y=c_{i})}{p_{tr}(y=c_{i})} \quad i=1...m \:.\label{eq:Weights_classCI}
\end{equation}

Let $\beta(x|y=c_{i})=\frac{p_{ts}(x|y=c_{i})}{p_{tr}(x|y=c_{i})}$ be the density ratio for class $c_{i}$ , and $\gamma(y=c_{i})=\frac{p_{ts}(y=c_{i})}{p_{tr}(y=c_{i})}$ be the ratio that reflects the changes of priors. Then, Eq. (\ref{eq:Weights_classCI}) can be written as
\begin{equation}
{w}_{c_i}=\beta(x|y=c_{i})\cdot\gamma(y=c_{i}) \:.
\end{equation}

As a result, we can induce the weights for all training samples in this class-wise manner, which reflects the changes of the joint distributions between the training and test data. Now, estimating the density ratio of joint distributions is divided into two sub-tasks: the estimation of the class-wise density ratio $\beta$ and the estimation of the prior-ratio $\gamma$.

However, the reality is that the test data do not have label information. In order to proceed with the class-wise matching, we propose an iterative procedure that alternates between estimating the class information for the test data and estimating new class-wise density ratios. The success of this iterative procedure greatly depends on two proposed components: (1) a soft distribution matching algorithm which incorporates the posteriors of the test data and, (2) an effective mutual information based stopping criterion. The details of the iterative procedures as well as the two components are explained in the rest of this section.

\subsection{Iterative Estimation Procedure.}

The iterative procedure proceeds as follows (Algorithm \ref{alg_ddr} shows the complete steps).

\begin{noindlist}
  \item The procedure begins with learning a classification model based on the training samples $\left\{ X_{tr},Y_{tr}\right\}$ whose weights are set to 1 ($w^{(0)}=1$) in the first iteration (Step 3).
  \item The classification model is then used to estimate the posteriors of the test data. It is noticeable that there is a distribution change between the training and test data, but we assume that the model can still give reasonable predictions for the posteriors of test samples \cite{Ben-David2010a} (Step 4).
  \item The class-wise density ratios are then estimated. Since the test data has extra information which is the posterior probabilities, we propose to utilize this extra information by extending the current density-ratio estimation techniques to incorporate the weighted test data. The details of this new method are explained in the soft matching section, named Soft Density-ratio (SoftDR) estimation (Step 5).
  \item The new prior ratios are estimated using an approach similar to that of Chang and Ng \cite{Chan2005}, in which we use the classifier's prediction results to update the priors (Step 6).
  \item Finally, the weight of training sample $w^{(t)}$ is updated (Step 7).
\end{noindlist}
The aforementioned steps are repeated until a stopping criterion is met.

\begin{algorithm}[t]
\caption{\label{alg_ddr}Discriminative Density-ratio Estimation}
 \textbf{Input:} $X_{tr}, Y_{tr}, X_{ts}$\\
 \textbf{Output:} $w(x), \space  x \in X_{tr}$\\
 \textbf{Steps:}
\begin{algorithmic}[1]
 \STATE{initialization: $w^{(0)} = 1;$\; $t = 0$\;}
 \WHILE{stop-criterion-not-met}
		\STATE{learn a model $\boldsymbol{\theta}^{(t)} = \text{Learn}(x_{tr},y_{tr}, w^{(t)})$\;}
		\STATE{predict $\hat{p}_{ts}^{(t+1)}(y|x\in X_{ts})=\boldsymbol{\theta}^{(t)}(x|x\in X_{ts})$\;}
		\STATE{estimate paired class: $\beta^{(t+1)}(x|y=c_{i})\Leftarrow \text{SoftDR}\left(X_{tr},Y_{tr},X_{ts},\hat{p}_{ts}^{(t+1)}(y|x\in X_{ts}), c_i\right)$\;}
		\STATE{estimate $\gamma^{(t+1)}(y=c_{i})=\frac{\hat{p}_{ts}^{(t+1)}(y=c_{i})}{p_{tr}(y=c_{i})}$\;}
		\STATE{update $w^{(t+1)} = \beta^{(t+1)} \cdot \gamma^{(t+1)}$\;}
		\STATE{$t = t+1$\;}
 \ENDWHILE
\end{algorithmic}
\end{algorithm}

\subsection{Soft Matching.}
\label{sec_softmatch}

Several density-ratio estimation methods have been proposed to estimate the importance (weights) of samples, in order to match two shifted distributions. The core concept of these methods relies on the kernel function to evaluate the similarity between samples, i.e. $k(x_{i},x_{j})$. The situation we face is that the test samples to be matched have soft decisions on the belongings to a class. An extension to the above kernel function can be used to utilize this information. 

Assume there are confidence scores (or probabilities) $w_{i}$ and $w_{j}$ associated with samples $x_{i}$ and $x_{j}$. Let $\phi(.)$ be the mapping function associated with the kernel. Then, the kernel function between two weighted samples can be calculated as
\begin{eqnarray}
k\left(\left\langle x_{i},w_{i}\right\rangle ,\left\langle x_{j},w_{j}\right\rangle \right) & = & \left[w_{i}\cdot\phi(x_{i})\right]^{T}\left[w_{j}\cdot\phi(x_{j})\right]\nonumber \\
 & = & w_{i}\cdot w_{j}\cdot\phi(x_{i})^{T}\cdot\phi(x_{j})\nonumber \\
 & = & w_{i}\cdot w_{j}\cdot k(x_{i},x_{j})\:.\label{eq:Kernel_weightedsamples}
\end{eqnarray}

Using the kernel of Eq. (\ref{eq:Kernel_weightedsamples}) with the original matching method allows the algorithm to perform soft Density-ratio (SoftDR) estimation. It is notable that the test sample confidence scores are induced by the posteriors and the training sample confidence scores are all
set to 1, because we have labels for the training data.

Without loss of generality, we illustrate the soft matching extension to Unconstrained Least-squares Importance Fitting (uLSIF) \cite{Kanamori2008} as an example.
In uLSIF, the density ratio is modeled as a linear combination of a series of basis functions as
\begin{equation}
\hat{\beta}(x)  =  \sum_{l=1}^{b}\alpha_{l}\varphi_{l}(x)\nonumber 
  =  \sum_{l=1}^{b}\alpha_{l}k(x,x_{l})\label{eq:beta_model}\:,
\end{equation}
where $x_{l}$ is a set of predefined reference points. 

Then, uLSIF learns the parameter $\left\{ \alpha_{l}\right\} _{l=1}^{b}$ by minimizing the squared loss of density-ratio function fitting. This leads to the following unconstrained optimization problem:
\begin{equation}
\min_{\left\{ \alpha_{l}\right\} _{l=1}^{b}}\left[\frac{1}{2}\sum_{l,l'=1}^{b}\alpha_{l}\alpha_{l'}\hat{S}_{l,l'}-\sum_{l=1}^{b}\alpha_{l}\hat{s}_{l}+\frac{\lambda}{2}\sum_{l=1}^{b}\alpha_{l}^{2}\right]\:,
\end{equation}
where
\begin{equation}
\hat{S}_{l,l'}:=\frac{1}{n_{tr}}\sum_{x_{i}\in\pi_{tr}}k(x_{i},x_{l})k(x_{i},x_{l'}),
\end{equation}
\begin{equation}
\hat{s}_{l}:=\frac{1}{n_{ts}}\sum_{x_{j}\in\pi_{ts}}k(x_{j},x_{l})\label{eq:uLSIF_h} \:.
\end{equation}

Because the training samples are given the label information, dividing them into groups according to their class labels means that the training samples have weights of 1. But the test samples are classified by a model in each iteration, whose output confidence values reflect their probability of belonging to a class. To reflect the uncertainty of test samples belonging to a class, we propose to add the soft matching ability to the uLSIF algorithm. Using the concept of weighted kernel functions (Eq. \ref{eq:Kernel_weightedsamples}), it can be observed that the objective function and $\hat{S}_{l,l'}$ are the same, except that $\hat{s}_{l}$ (Eq. \ref{eq:uLSIF_h}) needs to be modified as
\begin{equation}
\hat{s}_{l,c}:=\frac{1}{n_{ts}}\sum_{x_{j}\in\pi_{ts}}p(c|x_{j})\cdot k(x_{j},x_{l}) \:,
\end{equation}
where $p(c|x_{j})$ is the posterior of sample $x_{j}$ having a class label $c$.

Among the existing density-ratio estimators, uLSIF is robust and computationally efficient, and hence it is used in our experiments. For other density-ratio estimation methods, such as the Kernel Mean Matching (KMM) \cite{Huang2007} and the Kullback-Leibler Importance Estimation Procedure (KLIEP) \cite{Sugiyama2007}, the soft matching can also be implemented in a similar way by modifying the kernel functions as show in Eq. (\ref{eq:Kernel_weightedsamples}).

\subsection{Stopping Criterion.}

One na\"{i}ve criterion for stopping the algorithm is based on the convergence of the weights, i.e. $\left\Vert w^{(t+1)}-w^{(t)}\right\Vert \leq\epsilon$. However, we observed that this criterion only works when there is a clear separation between classes. For real datasets, this criterion will usually lead to a poor local solution. 
Instead, we propose to adopt the Mutual Information (MI) \cite{Wells1996}, as an indicator for a desirable location of the decision boundary. 

Given a test sample $x_{t}$, we define its posteriors using the current
model as a $m$ dimension vector (corresponding to $m$ classes) as $\hat{p}_{t}=\left[\hat{p}_{t1},\hat{p}_{t2},\ldots,\hat{p}_{tm}\right]^{T}$\:.

Then, the information entropy of this probability vector is defined as
\begin{equation}
H(\hat{p_{t}})=-\sum_{i=1}^{m}\hat{p}_{ti}ln(\hat{p}_{ti}) \:.
\end{equation}

MI between the test samples $X_{ts}$ and their estimated labels $\hat{Y}_{ts}$ using the model's output $\hat{p}_{ts}$ is defined as
\begin{equation}
\text{MI}\left(X_{ts},\left\langle \hat{Y}_{ts},\hat{p}_{ts}\right\rangle \right)=H(\hat{p}_{0})-\frac{1}{n_{ts}}\sum_{t=1}^{n_{ts}}H(\hat{p}_{t})\:,
\end{equation}
where $\hat{p}_{t}$ is the posterior vector for sample $x_{t}$, $\hat{p}_{0}$ is the class prior, and $H(.)$ is the information entropy.

\begin{table*}[tbh]
\begin{center}
\caption{\label{tab:The-distributions-Toy}The distributions of the training
and test data of the 2-class 4-cluster problem.}

\begin{tabular}{|c|c|c|c|}
\hline 
 &  & \textbf{\small{Prior}} & \textbf{\small{Likelihood}}\tabularnewline
\hline 
\hline 
\multirow{2}{*}{\small{$P_{tr}$}} & \small{class-1} & \small{0.5} & {\small{$0.9*\mathcal{N}\left(\left[\begin{array}{c}
1\\
5
\end{array}\right],\boldsymbol{I}\right)+0.1*\mathcal{N}\left(\left[\begin{array}{c}
4\\
5
\end{array}\right],\boldsymbol{I}\right)$}}\tabularnewline
\cline{2-4} 
 & \small{class-2} & \small{0.5} & {\small{$0.1*\mathcal{N}\left(\left[\begin{array}{c}
1\\
1
\end{array}\right],\boldsymbol{I}\right)+0.9*\mathcal{N}\left(\left[\begin{array}{c}
4\\
1
\end{array}\right],\boldsymbol{I}\right)$}}\tabularnewline
\hline 
\multirow{2}{*}{\small{$P_{ts}$}} & \small{class-1} & \small{0.6} & {\small{$0.5*\mathcal{N}\left(\left[\begin{array}{c}
1\\
5
\end{array}\right],\boldsymbol{I}\right)+0.5*\mathcal{N}\left(\left[\begin{array}{c}
4\\
5
\end{array}\right],\boldsymbol{I}\right)$}}\tabularnewline
\cline{2-4} 
 & \small{class-2} & \small{0.4} & {\small{$0.5*\mathcal{N}\left(\left[\begin{array}{c}
1\\
1
\end{array}\right],\boldsymbol{I}\right)+0.5*\mathcal{N}\left(\left[\begin{array}{c}
4\\
1
\end{array}\right],\boldsymbol{I}\right)$}}\tabularnewline
\hline 
\end{tabular}
\end{center}
\end{table*}

\begin{figure*}[]
\includegraphics[width=0.32 \textwidth]{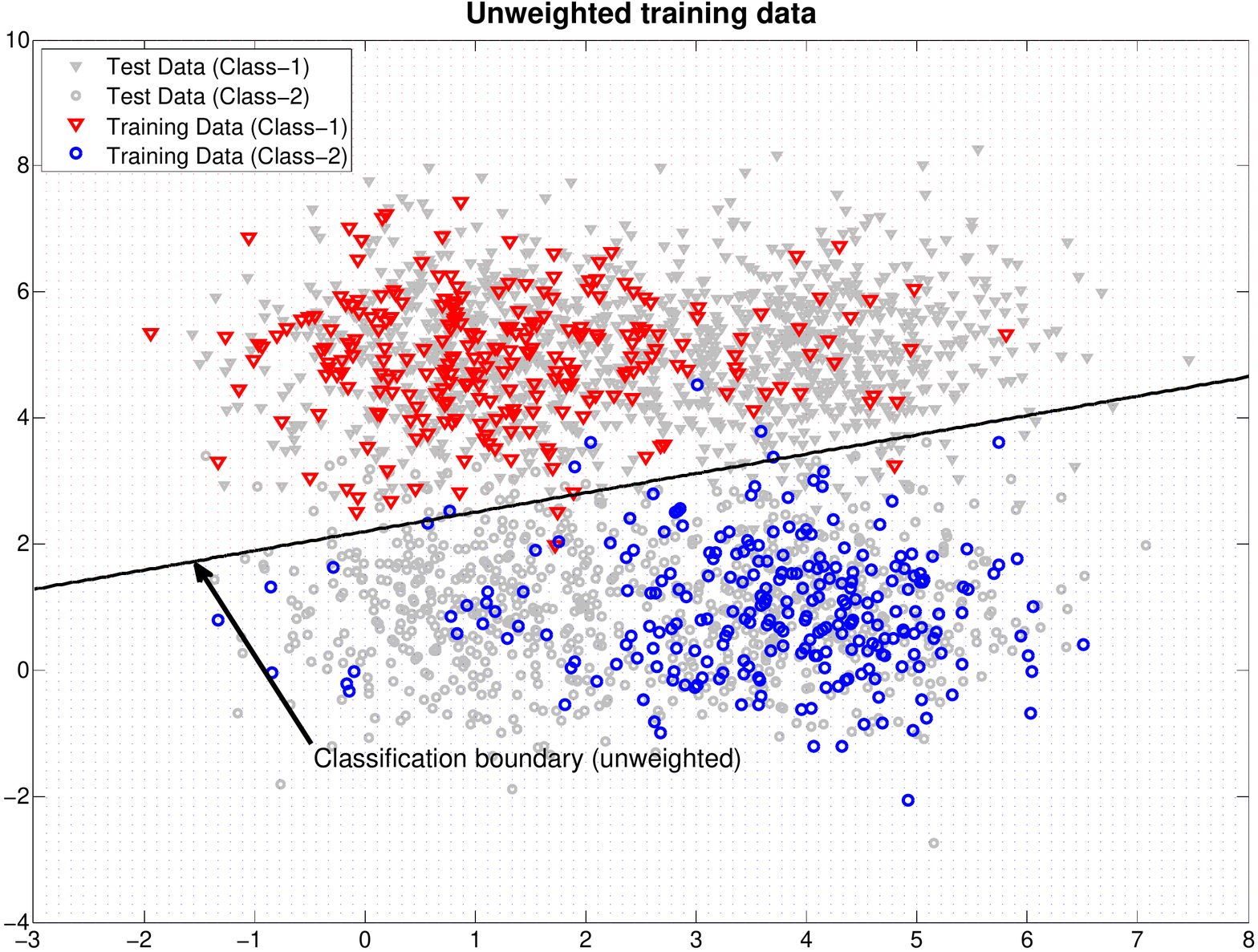}
\hspace{0.01 \textwidth}
\includegraphics[width=0.32 \textwidth]{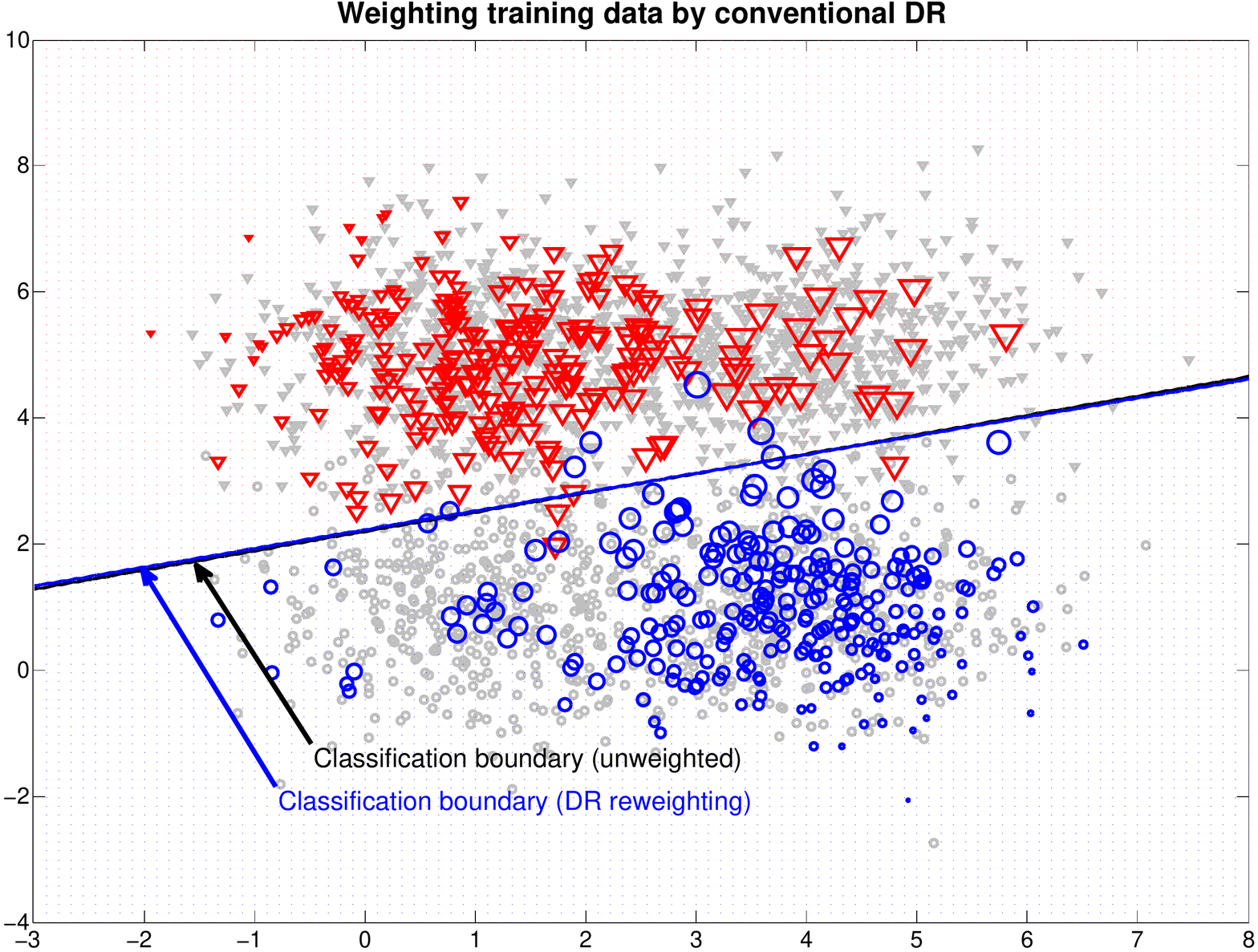}
\hspace{0.01 \textwidth}
\includegraphics[width=0.32 \textwidth]{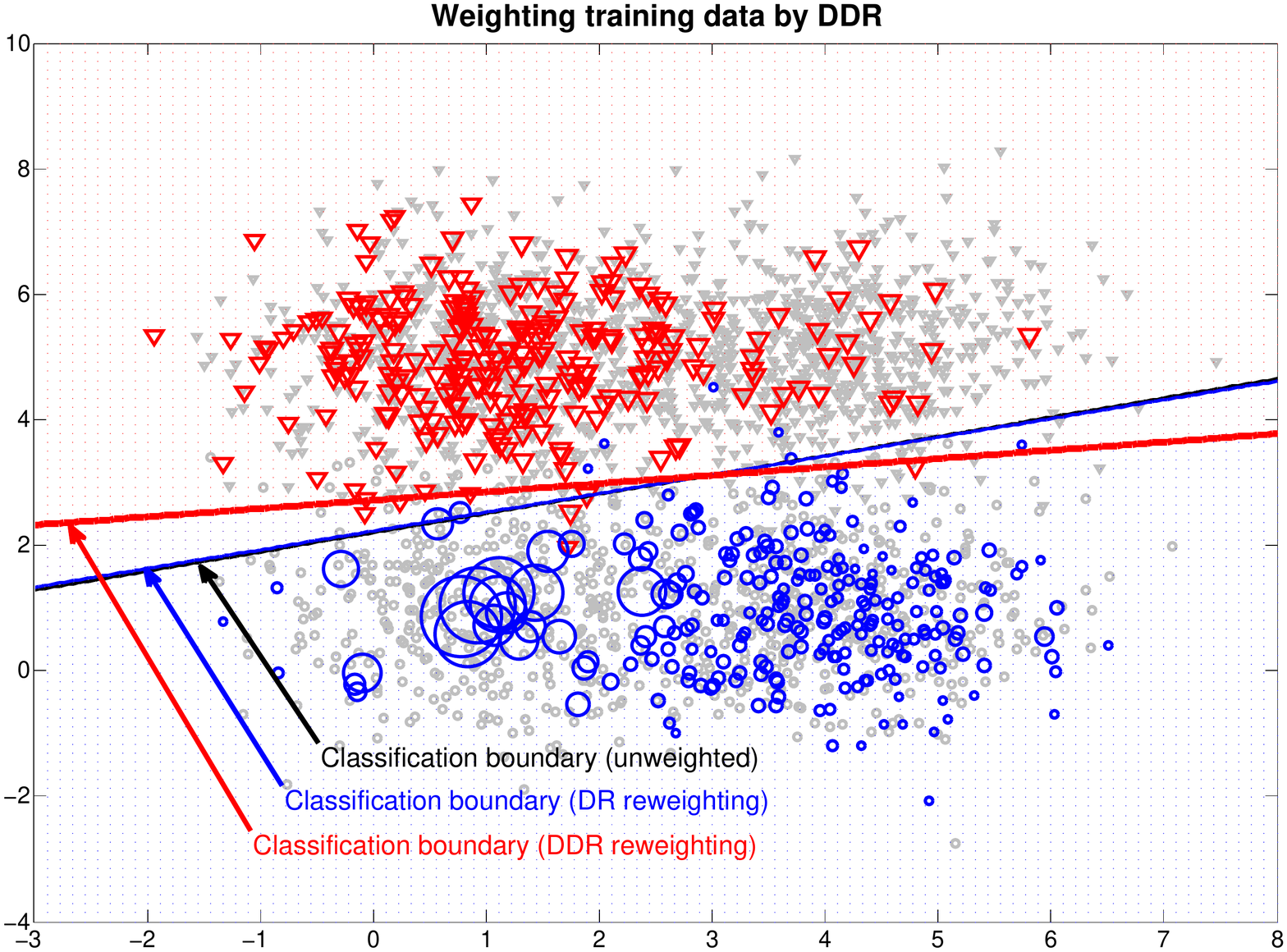}
\caption{\label{fig:ToyEx}Weighted training data and classification boundary: the unweighted training data (left), the conventional Density-ratio (DR) estimator (middle), and the Discriminative Density-ratio (DDR) estimator (right).}

\end{figure*}

Mutual information has been studied in the context of discriminative clustering \cite{Dhillon2003}, semi-supervised learning \cite{El-Yaniv2001} and domain adaptation \cite{Shi2012}. Maximizing this criterion implicitly means that the output of the current model has the least amount of confusing labels and the classification boundaries lie at sparse regions.
Facing the covariate shift scenarios and the unknown but drifted test distributions, we expect that this criterion could serve as a good indicator that can be utilized as an effective stopping condition.

\section{Experiments}
\label{exp}

In this section, we conduct three sets of experiments. The first one is on a synthetic 2-class
4-cluster data. The second experiment evaluates the sampling bias scenarios 
on different benchmark datasets. Following, a more challenging cross-dataset 
task is studied. 

\subsection{Synthetic Data.}

Our first experiment is designed with samples generated from 2-dimensional Gaussian mixture models, in which both the class priors and likelihoods exhibit changes. The 2-class 4-cluster distributions of the training and test
data are given as Table \ref{tab:The-distributions-Toy}.

Figure \ref{fig:ToyEx} illustrates the difference of importance estimation results between the unweighted approach, the conventional DR, and the proposed DDR methods, and their corresponding classification boundaries. We can observe that the conventional DR method assigns higher importance weights to the misclassified blue points because they lie in a dense region of test points (middle figure), while our proposed DDR method assigns small importance weights to these points and accordingly learns a much better decision boundary (right figure). The left figure clearly shows that classification using the unweighted approach is biased to training samples and leads to a suboptimal solution to the test data.

Table \ref{tab:Classification-accuracies-Toy} reports the numerical
classification results for various numbers of training samples using
the Naive Bayes Classifier. The number of test samples is fixed to
2000. Both DR and DDR are based on the density-ratio estimating method
of uLSIF. The results show that weighting the
covariate shifted data with the proposed DDR method performs consistently
better than the conventional DR method in term of the classification
accuracy.

As a reference line, we also report the experimental result
that is based on 5-fold cross validation on the test data, which
is the performance for test data without exposure to distribution changes (the column `Oracle-cvtest' in Table \ref{tab:Classification-accuracies-Toy}).

\begin{table*}[tbh]
\begin{center}
\caption{\label{tab:Classification-accuracies-Toy}Classification accuracies
over 30 runs for the 2-class 4-cluster data with variant number of
training samples. For each test case, the best-performing method other than the reference method `Oracle-cvtest'
is highlighted in bold (according to a $t$-test with 5\% significant level).}

\begin{tabular}{|c|c|c|c||c|}
\hline 
\small{$\boldsymbol{n_{tr}}$} & \textbf{\small{Unweighted}} & \textbf{\small{uLSIF}} & \textbf{\small{DDR-uLSIF}} & \textbf{\small{Oracle-cvtest}}\tabularnewline
\hline 
\hline 
\textbf{\small{100}} & {\small{0.9533\textpm{}0.0143}} & {\small{0.9587\textpm{}0.0182}} & \textbf{\small{0.9717\textpm{}0.0060}} & {\small{0.9770\textpm{}0.0035}}\tabularnewline
\hline 
\textbf{\small{200}} & {\small{0.9549\textpm{}0.0094}} & {\small{0.9655\textpm{}0.0095}} & \textbf{\small{0.9745\textpm{}0.0046}} & {\small{0.9778\textpm{}0.0036}}\tabularnewline
\hline 
\textbf{\small{300}} & {\small{0.9545\textpm{}0.0101}} & {\small{0.9645\textpm{}0.0085}} & \textbf{\small{0.9735\textpm{}0.0032}} & {\small{0.9771\textpm{}0.0026}}\tabularnewline
\hline 
\textbf{\small{400}} & {\small{0.9553\textpm{}0.0072}} & {\small{0.9641\textpm{}0.0087}} & \textbf{\small{0.9739\textpm{}0.0042}} & {\small{0.9762\textpm{}0.0035}}\tabularnewline
\hline 
\textbf{\small{500}} & {\small{0.9585\textpm{}0.0065}} & {\small{0.9676\textpm{}0.0064}} & \textbf{\small{0.9736\textpm{}0.0048}} & {\small{0.9770\textpm{}0.0030}}\tabularnewline
\hline 
\textbf{\small{1000}} & {\small{0.9596\textpm{}0.0056}} & {\small{0.9681\textpm{}0.0052}} & \textbf{\small{0.9737\textpm{}0.0046}} & {\small{0.9776\textpm{}0.0031}}\tabularnewline
\hline 
\end{tabular}
\end{center}
\end{table*}

\subsection{Biased Sampling.}

Further, we evaluate our proposed DDR method on a set of benchmark datasets. The datasets `GermanCredit', `DelveSplice', `Ionosphere', `Australian', `BreastCancer', `Diabete' are from the UCI Machine Learning Repository\footnote{\url{http://archive.ics.uci.edu/ml/datasets.html}}. The datasets `USPS' and `MNIST' are from the LibSVM data collection\footnote{\url{http://www.csie.ntu.edu.tw/~cjlin/libsvmtools/datasets/}}.

The covariate shift classification tasks are formulated by splitting the training and test data with a deliberate biased
sampling selection procedure (following the setup of \cite{Cortes2008}).
In all experiments, before any further processing,
all the data are normalized to the range $[-1,1]^{d}$.  
Then, the half of data are uniformly sampled to form the testing
section. And, the rest of data are sub-sampled to form the biased
training set with the probability of $P\left(s=1|x\right)=\frac{e^{v}}{1+e^{v}}$,
where $s=1$ means the sample is included in the training
set, and $v=\frac{4\boldsymbol{\omega^{t}}(x-\overline{x})}{\sigma_{\boldsymbol{\omega^{t}}(x-\overline{x})}}$.
$\boldsymbol{\omega} \in {R}^{d}$ is a projection vector randomly
chosen from $\left[-1,1\right]^{d}$. For each run of experiment, ten
vectors of $\boldsymbol{\omega}$ are randomly generated and we select the vector $\boldsymbol{\omega}$
which maximizes the difference between the unweighted method and the
weighted method with ideal sampling weights. 

We employ the uLSIF method in our experiments because of its superiority in speed and numerical stability. 
The classifier we used is Importance-Weighted
Least-Squares Probabilistic Classifier (IWLSPC) \cite{Hachiya2012}.
The number of kernel basis functions is set to 100 by random sampling
from the test data. The other hyper parameters (the kernel width $\sigma$ and
regularization parameter $\lambda$) are chosen by 5-fold Importance Weighted
Cross-Validation (IWCV) \cite{Sugiyama2007}.

We evaluate the performance of our DDR method by comparing with the
conventional density-ratio estimation method using the exact same settings.
The classification results using the model learned from the unweighted training data
are included as the baseline. Because of the deliberate biased
sampling selection procedure, we know the probability of each sample being included into the training section is $P\left(s=1|x\right)$. Therefore, the perfect sample importance is known as the reciprocal of being selected, i.e., $\text{imp} = \frac{1}{P\left(s=1|x\right)}$. We report results of using this oracle importance weights as the reference (the column `Oracle-imp' in Table \ref{tab:DigitsRecognition}).

All experiments are repeated 30 times with different training-test data splits. The significance of the improvement in classification accuracy is tested using a $t$-test at a significance level of 5\%.
The results are summarized in Table \ref{tab:DigitsRecognition}. It can be observed that the proposed DDR approach outperforms the unweighted method and the conventional density-ratio estimator in almost all cases. There are 4 out of 10 cases where the accuracies are improved by more than 10\%. 

\begin{table*}[tbh]
\begin{center}
\caption{\label{tab:DigitsRecognition}Biased sampling on benchmark datasets: Classification accuracies over 30 runs. For each dataset, the best-performing method is highlighted in bold (according to a $t$-test with 5\% significant level, the setup of `Oracle-imp' is a reference and not involved to comparison).
}

\begin{tabular}{|c|c|c|c||c|}
\hline 
\textbf{\small{Dataset}} & \textbf{\small{Unweighted}} & \textbf{\small{uLSIF}} & \textbf{\small{DDR-uLSIF}} & \textbf{\small{Oracle-imp}} \tabularnewline
\hline 
\textbf{\small{GermanCredit}} & \small{0.6970\textpm{}0.0383}  & \small{0.6888\textpm{}0.0554}  & \textbf{\small{0.7013\textpm{}0.0130}}  & \small{0.6935\textpm{}0.0495}   \tabularnewline
\hline 
\textbf{\small{DelveSplice}} & \small{0.5527\textpm{}0.0601}  & \small{0.5797\textpm{}0.0797}  & \textbf{\small{0.6679\textpm{}0.1184}}  & \small{0.6336\textpm{}0.0874}   \tabularnewline
\hline 
\textbf{\small{Ionosphere}} & \small{0.6818\textpm{}0.0565}  & \small{0.6759\textpm{}0.0588}  & \textbf{\small{0.6979\textpm{}0.0666}}  & \small{0.7483\textpm{}0.0866}   \tabularnewline
\hline 
\textbf{\small{Australian}} & \small{0.8121\textpm{}0.0298}  & \small{0.8187\textpm{}0.0450}  & \textbf{\small{0.8319\textpm{}0.0275}}  & \small{0.8342\textpm{}0.0272}   \tabularnewline
\hline 
\textbf{\small{BreastCancer}} & \small{0.8189\textpm{}0.1498}  & \small{0.7963\textpm{}0.1379}  & \textbf{\small{0.9219\textpm{}0.1107}}  & \small{0.8942\textpm{}0.1361}   \tabularnewline
\hline 
\textbf{\small{Diabete}} & \textbf{\small{0.7372\textpm{}0.0245}} & \textbf{\small{0.7346\textpm{}0.0273}} & \textbf{\small{0.7286\textpm{}0.0279}} & \small{0.7149\textpm{}0.0378} \tabularnewline
\hline 
\textbf{\small{USPS5v6}} & {\small{0.9581\textpm{}0.0081}} & {\small{0.9508\textpm{}0.0297}} & \textbf{\small{0.9747\textpm{}0.0062}} & \small{0.9689\textpm{}0.0163}  \tabularnewline
\hline 
\textbf{\small{USPS3v8}} & {\small{0.6262\textpm{}0.0623}} & {\small{0.7443\textpm{}0.1549}} & \textbf{\small{0.9283\textpm{}0.0813}} & \small{0.7861\textpm{}0.1561}  \tabularnewline
\hline 
\textbf{\small{MNIST5v6}} & {\small{0.7979\textpm{}0.1978}} & {\small{0.8888\textpm{}0.1353}} & \textbf{\small{0.9477\textpm{}0.0100}} & \small{0.9124\textpm{}0.1384}  \tabularnewline
\hline 
\textbf{\small{MNIST3v8}} & {\small{0.5591\textpm{}0.1079}} & {\small{0.5725\textpm{}0.1228}} & \textbf{\small{0.7936\textpm{}0.1640}} & \small{0.6809\textpm{}0.1819}  \tabularnewline
\hline 
\hline 
\end{tabular}
\end{center}
\end{table*}

\subsection{Cross-dataset Tasks.}

Training a model with samples from one dataset and adapting the model to another dataset which is collected at different conditions, is usually seen as a very challenging problem. We evaluate our DDR approach in the cross-dataset classification task using the two handwritten digits recognition datasets: USPS and MNIST.
The USPS dataset contains 9,298 handwritten digit images with the size $16 \times 16$. The MNIST dataset has a total of 70,000 handwritten digit images (the first 20,000 samples are used in our experiment). The size of each image is $28 \times 28$. 

Because the two datasets have different image sizes and intensity levels, a preprocessing step is applied first as: (1) resize the image size of MNIST from $28 \times 28$ into the same size of USPS, $16 \times 16$; (2) normalize the feature (intensity of pixel) into the range of $[-1,1]$. Then, we conduct two scenarios of experiments: one using USPS for training and MNIST for testing, the other using MNIST for training and USPS for testing. The classification method being used is SVM with linear kernels \cite{Chang2011a}. The parameter $c$ in SVM is a trade-off between model generalization and training error, and its value is chosen using 5-fold importance weighted cross-validation.

Table \ref{tab:CrossDataset} presents the average and standard deviations of the classification accuracies of 30 runs. Each run is based on randomly selecting 90\% of training samples and test samples from the datasets. The report results show that the DDR method can significantly boost recognition accuracies. Compared to the conventional DR approach, for the scenario ``USPS to MNIST" 7 out of 10 test cases achieve an improvement in accuracy of 2\% to 7\%. For the scenario ``MNIST to USPS", 8 out of 10 test cases gain an improvement in accuracy of 2\% to 15\%.

\begin{table*}[]
\begin{center}
\caption{\label{tab:CrossDataset}Cross-dataset tasks: classification accuracies
over 30 runs on the USPS and MNIST datasets. For each test case, the best-performing method is highlighted in bold according to a $t$-test with 5\% significant level. The italic means that it records the best average accuracy but not statistically significant.}

\begin{tabular}{|c|c|c|c||c|c|c|}
\hline 
 & \multicolumn{3}{c|}{\textbf{\small{USPS to MNIST}}} & \multicolumn{3}{c|}{\textbf{\small{MNIST to USPS}}}\tabularnewline
\hline 
\textbf{\small{Test Case}} & \textbf{\small{Unweighted}} & \textbf{\small{uLSIF}} & \textbf{\small{DDR-uLSIF}} & \textbf{\small{Unweighted}} & \textbf{\small{uLSIF}} & \textbf{\small{DDR-uLSIF}}\tabularnewline
\hline 
\textbf{\small{0 vs 1}} & \small{0.8695\textpm{}0.0619}  & \small{0.8751\textpm{}0.0627}  & \small{\textbf{0.9337\textpm{}0.0682}}  & \small{0.9455\textpm{}0.0098} & \small{0.9480\textpm{}0.0087}  & \small{\textbf{0.9697\textpm{}0.0129}} \tabularnewline
\hline 
\textbf{\small{1 vs 2}} & \small{0.5947\textpm{}0.0280}  & \small{0.5910\textpm{}0.0277}  & \small{\textit{0.6008\textpm{}0.0247}}  & \small{0.9114\textpm{}0.0270}  & \small{0.9094\textpm{}0.0405}  & \small{\textbf{0.9245\textpm{}0.0301}} \tabularnewline
\hline 
\textbf{\small{2 vs 3}} & \small{0.7200\textpm{}0.0756}  & \small{0.6958\textpm{}0.0997}  & \small{\textbf{0.7688\textpm{}0.0580}}  & \small{0.6432\textpm{}0.0393}  & \small{0.6470\textpm{}0.0309}  & \small{0.6412\textpm{}0.0334} \tabularnewline
\hline 
\textbf{\small{3 vs 4}} & \small{0.7991\textpm{}0.0421}  & \small{0.8083\textpm{}0.0479}  & \small{\textbf{0.8226\textpm{}0.0323}}  & \small{0.7878\textpm{}0.0390}  & \small{0.7859\textpm{}0.0332}  & \small{\textbf{0.8129\textpm{}0.0258}} \tabularnewline
\hline 
\textbf{\small{4 vs 5}} & \small{0.6373\textpm{}0.0624}  & \small{0.7046\textpm{}0.0471}  & \small{\textbf{0.7279\textpm{}0.0656}}  & \small{0.8322\textpm{}0.0317}  & \small{0.8440\textpm{}0.0477}  & \small{\textbf{0.8794\textpm{}0.0173}} \tabularnewline
\hline 
\textbf{\small{5 vs 6}} & \small{0.5644\textpm{}0.0593}  & \small{0.5336\textpm{}0.0385}  & \small{\textbf{0.5861\textpm{}0.0705}} & \small{0.5901\textpm{}0.0280}  & \small{0.5778\textpm{}0.0306}  & \small{\textbf{0.6787\textpm{}0.0507}} \tabularnewline
\hline 
\textbf{\small{6 vs 7}} & \small{0.6025\textpm{}0.0820}  & \small{0.6041\textpm{}0.0807}  & \small{0.6039\textpm{}0.0813}  & \small{0.5105\textpm{}0.0276}  & \small{0.5106\textpm{}0.0280}  & \small{\textit{0.5150\textpm{}0.0320}} \tabularnewline
\hline 
\textbf{\small{7 vs 8}} & \small{0.6290\textpm{}0.0631}  & \small{0.6164\textpm{}0.0642}  & \small{\textbf{0.6352\textpm{}0.0697}}  & \small{0.6240\textpm{}0.0407}  & \small{0.6220\textpm{}0.0484}  & \small{\textbf{0.7775\textpm{}0.0440}} \tabularnewline
\hline 
\textbf{\small{8 vs 9}} & \small{0.6977\textpm{}0.0783}  & \small{0.7376\textpm{}0.0825}  & \small{\textbf{0.8066\textpm{}0.0826}}  & \small{0.8255\textpm{}0.0622}  & \small{0.7866\textpm{}0.0540}  & \small{\textbf{0.8340\textpm{}0.0509}} \tabularnewline
\hline 
\textbf{\small{9 vs 0}} & \small{0.9092\textpm{}0.0314} & \small{0.9133\textpm{}0.0302} & \small{\textit{0.9142\textpm{}0.0287}} & \small{0.6482\textpm{}0.1067} & \small{0.6775\textpm{}0.1131} & \small{\textbf{0.8099\textpm{}0.1087}}\tabularnewline
\hline 
\end{tabular}
\end{center}

\end{table*}

\section{Conclusion}
\label{con}

This paper proposes a novel algorithm for covariate shift classification problems
which estimates the density-ratio in a discriminative manner. Instead of matching the marginal distributions without paying attention to the separations among classes, our proposed method estimates density ratio between joint distributions in a class-wise manner. Therefore, our method allows relaxing the strong assumption of covariate shift and
preserves the separation between classes while minimizing the distribution discrepancy between the training and test data.
In order to proceed with the class-wise matching, the proposed algorithm deploys an iterative procedure that alternates between estimating the class information for the test data and estimating new class-wise density ratios. Two modules contribute to the success of the proposed DDR method. One is the soft matching algorithm which extends current density-ratio estimation algorithms to incorporate sample posterior. Another important component is the employment of the  mutual information as an indicator for stopping the iterative procedure. Experiments on synthetic and benchmark data confirm the superiority of the proposed  algorithm.

Although our method focused on the covariate shift adaptation problem, we vision that the concept of discriminative distribution
matching is also useful to other scenarios of transfer learning.

\balance
\bibliographystyle{unsrt}
\bibliography{DDR_ReferenceList}

\end{document}